\documentclass[a4paper,twoside,french]{article}
\usepackage{rfia2000}
\usepackage[T1]{fontenc}
\usepackage{babel}
\usepackage{times}
\usepackage{bbold}
\usepackage{amssymb}
\usepackage{hyperref}
\usepackage{times}
\usepackage{amsmath,graphicx}
\usepackage{lscape}

\usepackage{soul} 

\usepackage{adjustbox} 

\usepackage{xcolor}

\usepackage[title]{appendix}

\usepackage{comment}

\def\identity{{\mathbb{1}}}
\def\loss{{\mathcal{L}}}
\def\ttlfaces{{\textit{TLL}$_{faces}$}}
\def\ttlobj{{\textit{TLL}$_{obj}$}}
\def\@fnsymbol#1{\ensuremath{\ifcase#1\or *\or \dagger\or \ddagger\or
   \mathsection\or \mathparagraph\or \|\or **\or \dagger\dagger
   \or \ddagger\ddagger \else\@ctrerr\fi}}
\newcommand{\ssymbol}[1]{^{\@fnsymbol{#1}}}

\begin{document}
%%%%%Pas de date
\date{}
%%%%% Titre gras 14 points
\title{\Large\bf Learning an adaptation function to assess image visual similarities}
\author{\begin{tabular}[t]{c@{\extracolsep{3em}}c@{\extracolsep{3em}}c@{\extracolsep{3em}}c@{\extracolsep{3em}}c}
O. Risser-Maroix${}^{1, 2}$  & A. Marzouki${}^1$ & H. Djeghim${}^1$ & C. Kurtz${}^1$ &N. Loménie${}^1$ \\
\end{tabular}
{} \\
 \\
${}^1$        LIPADE, Université de Paris -- Paris, France  
{} \\
 \\
${}^2$\url{orissermaroix@gmail.com}
}
\maketitle
%%%%  Pas de numérotation sur la page de titre
\thispagestyle{empty}

\subsection*{Abstract}
{\em
Human perception is routinely assessing the similarity between images, both for decision making and creative thinking. 
But the underlying cognitive process is not really well understood yet, hence difficult to be mimicked by computer vision systems.
State-of-the-art approaches using deep architectures are often based on the comparison of images described as feature vectors learned for image categorization task. 
As a consequence, such features are powerful to compare semantically related images but not really efficient to compare images visually similar but semantically unrelated.
Inspired by previous works on neural features adaptation to psycho-cognitive representations, we focus here on the specific task of learning visual image similarities when analogy matters.
We propose 
{to compare different supervised, semi-supervised and self-supervised networks, pre-trained on distinct scales and contents datasets (such as 
ImageNet-21k, ImageNet-1K or VGGFace2)
to conclude which model may be the best to approximate}
the visual cortex and learn only an adaptation function corresponding to the approximation of the the primate IT cortex through the metric learning framework. 
Our experiments conducted on the Totally Looks Like image dataset highlight the interest of our method, by increasing the retrieval scores {of the best model} @1 by 2.25×. 
This research work was recently accepted for publication at the ICIP 2021 international conference \cite{risser_maroix2021}. In this new article, we expand on this previous work by using and comparing new pre-trained feature extractors on other datasets.
}
\subsection*{Keywords}
Visual Similarity, Features Adaptation, Image Retrieval, Analogies

\section{Motivation}
\label{sec:intro}

Analogies are constantly employed by humans to find connections / similarities between images, when learning concepts or for creative purposes.
Our perception being extremely complex to model, it remains difficult to imitate by machines.
However, capturing image similarities is a cornerstone in various computer vision tasks, such as retrieval, classification, spotting, etc. 
This task is challenging since an image has many more interpretations than its textual description and the sought similarity may depend on a hidden intention.

State-of-the-art approaches often rely on the comparison of images described as vectors of features learned via convolutional neural networks (CNNs) with objectives close to classification task. 
Since such features are generally learned for image categorization, they are biased by the semantics of the decision process of the classifiers.
In this article, we focus on the task of learning image visual similarities and we involve this in an image retrieval context.
Inspired by previous works on neural features adaptation to psycho-cognitive representations, we propose a way to learn a perceptual similarity function between images that share visual connections but that are semantically unrelated.
To our knowledge, we are among the first to propose a methodology to face this issue of understanding the underlying psycho-visual process of matching images of different natures. 

\section{Background}

In human perception, the notion of similarity between concepts or even images has been studied for a while and remains extremely difficult to define \cite{Tversky1978StudiesOS, rogowitz1998perceptual, Tirilly2012, Liu2001, Mur2013}.
During the last decade, scientists from cognitive and computer sciences started to analyze the differences between human and machine perception and to investigate on how modern neural architectures could help to capture human judgments of similarity; 
such a similarity can be either guided by general concepts \cite{Peterson2018, ZhengPBH19, Hebart2020} or performed by strictly visual correspondences \cite{Kubilius2016, Pramod2016, Sadovnik_2018_CVPR_Workshops, zhang2018perceptual, German2020}.
As already stated by \cite{Rosenfeld2018}, strategies based on visual stimuli yield good results with very simple images (with well segmented background) \cite{Pramod2016} or when focusing on a narrow class \cite{Peterson2018, Sadovnik_2018_CVPR_Workshops,hamilton2020conditional} (such as only faces, animals, arts).

\begin{figure}[t]
%\begin{minipage}[b]{1.0\linewidth}
  \centering
  \centerline{\includegraphics[width=7.5cm]{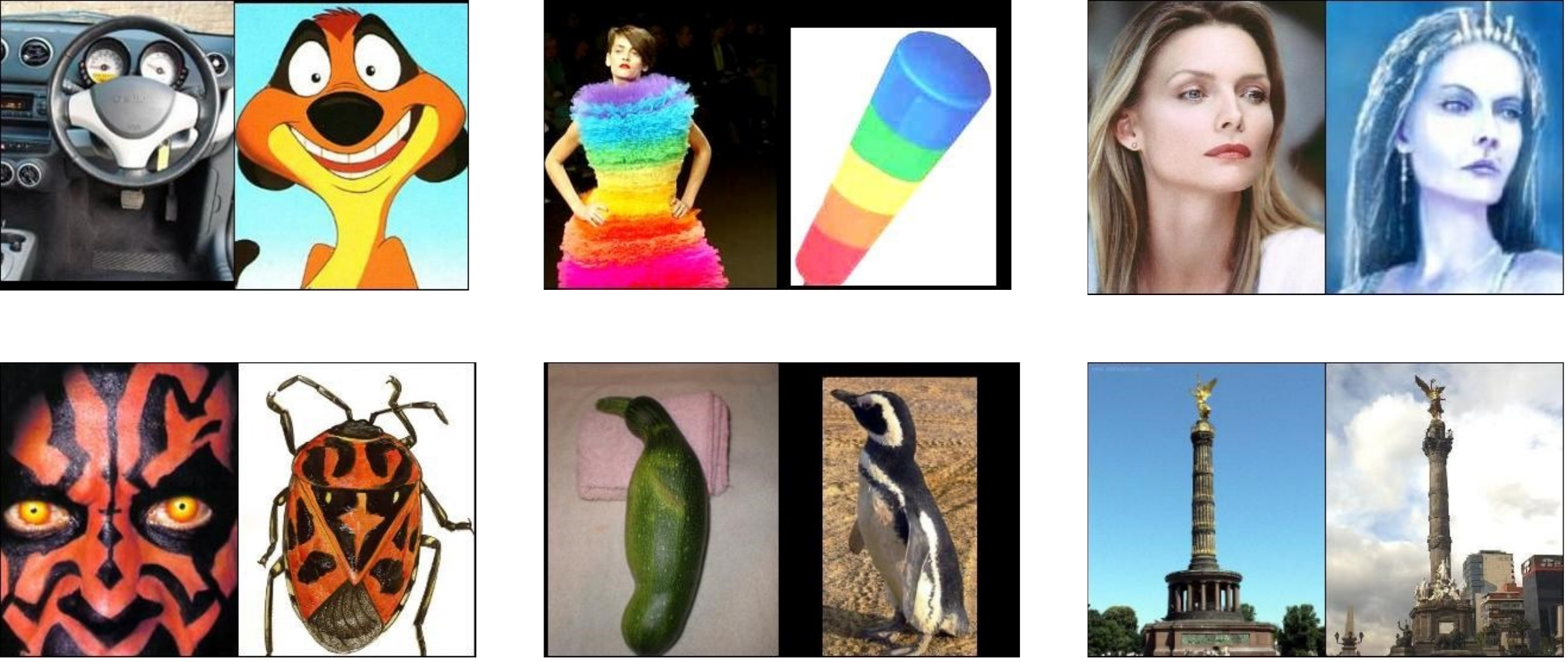}}
  %Fig.\ref{fig:ttl_dataset} -- \textit{Examples of images pairs in the "Totally Looks Like dataset". Similarity can be based on the color, shape, texture, layout, facial similarity, etc.}\medskip
  \caption{Examples of image pairs from the "Totally Looks Like" dataset. 
  Similarity can be based on the color, shape, texture, layout, facial similarity, etc.}
%\end{minipage}
\label{fig:ttl_dataset}
\end{figure}

The \textit{Totally Looks Like} image dataset (denoted as \textit{TLL}) is an interesting example of this research area \cite{Rosenfeld2018}.
\textit{TLL} is composed of 6k+ image pairs resulting from human propositions, where similarity between images is not based on semantic categorization (as this is the case in most classical image retrieval datasets) but only based on visual clues derived from the image contents.
The images belong to different domains such as photography, cartoons, sketches, logos, etc. making the task even harder but more representative of human ability to make connections between semantically unrelated objects (some illustrative samples are provided in Fig.~\ref{fig:ttl_dataset}). 
From this dataset, one can note that the process of human visual similarity mixes multiple different levels of analysis ranging from color, texture, to shape, layout, etc. in which context, cultural aspect and possibly humor and irony can play an important role.

Ones could argue that similarity is a too subjective and an ill-posed problem.
But given a dataset of image pairs such as \textit{TLL}, authors from \cite{Rosenfeld2018} conducted human experiments %through Amazon Mechanical Turk 
and found that when other image candidates were proposed to form pairs, humans remained consistent in their choices and selected invariably the right target image to make the pair. 
%The same observations were also made by former studies \cite{Liu2001, Tirilly2012}. %(ANY BIAS with the WORKERS ANNOTATING?)
Authors from \cite{Liu2001} found that results in different visual studies are highly influenced by the task requirements; in our case the task is defined by the visual similarity without taking into account the semantic similarity.
In addition, \cite{Tirilly2012} showed that, for the task of grading the similarity between images, the quantitative analysis of the similarity scores reported by subjects reached a consensus. 

Authors from \cite{Rosenfeld2018} tried to reproduce the human similarity judgments from a given image pairs dataset with different neural architectures and, as mentioned earlier, they found that the learned convolutional features were not adapted for this task.
Later \cite{Rosenfeld2019} improved the similarity scores by using even more descriptors crafted for different uses (color only, shape only, etc.), and by using the right descriptor for each pair as oracle. 
None of \cite{Rosenfeld2018, Rosenfeld2019} worked on features adaptation, neither learned a model on a part of the dataset, making us the first to propose a baseline for this task.
As comparative work, starting from the observation that traditional metrics (L2, PSNR, etc.) disagree with human judgments, authors from \cite{zhang2018perceptual} already learned a low-level perceptual measure of similarity from image patches affected by distortions.
The poor results obtained by \cite{Rosenfeld2019} on \textit{TLL} by using the low level perceptual learned similarity from \cite{zhang2018perceptual}, as well as with higher level semantic features confirm the previous study of \cite{Mur2013} stating that human visual similarity was not based in the visual cortex but may be the result of processing done in the primate inferior temporal (IT) cortex. 
In our case, we hypothesize that even if neural networks are learned to be robust to cases where images are semantically similar but visually dissimilar, they still carry useful (and reusable) information about texture, shapes, etc. 
When dealing with relatively small datasets such as \textit{TLL}, based on a strategy originally introduced in  \cite{Mur2013}, we propose to use different layers of pre-trained networks 
with the opposite objective of categorization as a rough approximation of the visual cortex and learn only an adaptation function corresponding to the approximation of the the primate IT cortex through the \textit{metric learning} framework.

\textit{Similarity learning} is closely related to \textit{distance metric learning} where the goal is to learn a distance function over objects that measures how similar two objects are.
In our case, we look for a model able to bring closer each training image pair, while moving away all other images that would form a less good pair than the ground-truth one.
For each image, we only have a single solution which makes our problem close to one-shot learning where metric learning has been considered \cite{koch2015siamese} by using \textit{Siamese networks} \cite{chopra2005learning}.
Another wildely used architecture in deep metric learning is the \textit{triplet network} \cite{hoffer2015deep} where for each positive pair, a negative image is also provided to learn simultaneously to gather the positive images while increasing the distance to the negative image.
In a similar fashion, \cite{Hebart2020} was able to provide powerful representations capturing human behavior by asking participants to select the \textit{one-odd-out} from a triplet of images and thus learning a model to mimic it.
It has been shown that those architectures suffer from sampling issues that could be partially solved with complex mining strategies \cite{Schroff2015, Wu2017}.

For the sake of simplicity, we rely here on the hypothesis that each image in a pair has a stronger connection than with any other image in the dataset.
We then learn a function able to bring closer the right image to the query than any other image.
By doing so, we aim to bypass the sampling issue and use a generalization of the triplet loss similar to the \textit{N-Loss} \cite{Sohn2016ImprovedDM}.
Inspired from previous works in cognitive sciences, we propose the first baseline for the task of visual similarity between images without taking into account semantic similarity.

This research work was recently accepted for publication at the ICIP 2021 international conference. 
In this new article, we expand on this previous work by using and comparing new pre-trained feature extractors on other datasets.

\section{Learning image visual similarities}
\label{sec:method}

\begin{figure}[t]
    \begin{minipage}[b]{1.0\linewidth}
      \centering
     % \centerline{\includegraphics[width=8.5cm]{imgs/pipeline_simple}}%learning_pipeline
      \centerline{\includegraphics[width=8.5cm]{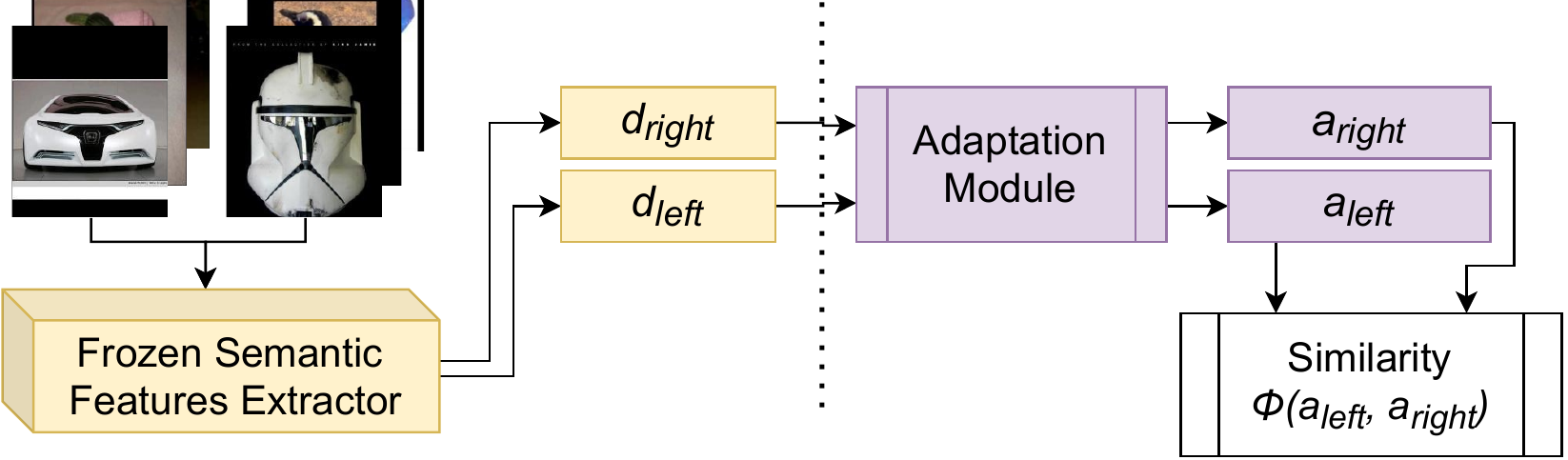}}%learning_pipeline
    %  \vspace{2.0cm}
      %\centerline{(a) Result 1}\medskip
    \end{minipage}

    \begin{minipage}[b]{.4\linewidth}
      \centering
    %  \centerline{\includegraphics[width=4.0cm]{image3}}
    %  \vspace{1.5cm}
      \centerline{(a) Image representations}\medskip
    \end{minipage}
    %\hfill 
    ~~~~~~~~~
    \begin{minipage}[b]{0.4\linewidth}
      \centering
    %  \centerline{\includegraphics[width=4.0cm]{image4}}
    %  \vspace{1.5cm}
      \centerline{(b) Adaptation function learning}\medskip
    \end{minipage}
    \caption{Pipeline to learn an adaptation function able to compute visual similarity from image pairs. %\comment{j'ai changé la figure, ça va ? c'est lisible et comprehensible ?}
    }
    \label{fig:pipeline}
\end{figure}

We propose to learn an adaptation function able to compute visual similarity from image pairs and to involve it in an image retrieval task (Fig.~\ref{fig:pipeline}).
Inspired by \cite{Mur2013}, the features extraction part (Fig.\ref{fig:pipeline}~(a)) could be interpreted as the visual cortex while the learned adaptation matrix and the measure deciding on the visual similarity between two images (Fig.\ref{fig:pipeline}~(b)) can be viewed as the primate IT cortex.

\subsection{Image retrieval task}

We assume that we have an image dataset, structured as pairs of visually similar images (considered as the ground-truth (GT)).
We consider each pair as the juxtaposition of a \textit{left} image and and a \textit{right} image, leading to two sets of images.
As in \cite{Rosenfeld2018}, we formulate this problem as an image retrieval task.
For each query in the left set of images, we rank all the images in the set of right, according to a given similarity measure $\phi(\cdot, \cdot)$, and reciprocally.
From this ranking, we want that for each query, the best returned candidate is the one expected by the GT pair.

We know from \cite{tversky1977features} that asymmetry in human judgment of similarity is important.
For example, in Fig.~\ref{fig:ttl_dataset}, most humans will think that the zucchini looks like the penguin rather than the penguin looks like the zucchini.
In our case, as we do not have the direction information of which image from left or right is looking to the other one, we cannot use an asymmetric function such as the \textit{Tversky Ratio Model} \cite{tversky1977features} and $\phi(\cdot, \cdot)$ is here a simple \textit{cosine similarity}.
However, we will embed this asymmetry in the evaluation function.

\subsection{Image representation}
{As pre-processing, we extract visual features for each image (right, left) from pre-trained networks. 
Different layers were used for some architectures based on Residual Networks while only the last layer was adopted for more complex ones due to the difficulty
{to localize} 
{the middle layers} (
FaceNet and transfomers).
{For more details, please refer to Table~\ref{table_scores} in the annex.}
}
\newline

We reduce the feature maps of each layer to a simple vector with the same dimension as the number of channels, by averaging features maps on the spatial dimensions followed by \textit{L2} normalization.
By doing so, each image is represented as the concatenation of features extracted from different layers to capture information, from low to high levels.
To avoid overfitting, we reduce the number of dimension of the original vectors (up to 15k+) with a PCA.
Left and right vectors are respectively named $d_{left}$, $d_{right}$ and form the image representations.

This pre-processing step (Fig.~\ref{fig:pipeline}~(a)) is done once before training and can be interpreted as a rough approximation of visual features extracted by the human visual cortex.
% \olivier
{
    Indeed, the object recognition and visual similarity assessment is done in the Ventral Stream which is composed of the Visual Cortex and the Primate IT Cortex \cite{lindsay2020convolutional}. The Visual Cortex is decomposed into  sub-regions (V1-to-V4); V4 is particularly known to respond to orientation, color, disparity and simple shapes and it is directly connected to the Primate IT Cortex. Since the different features V4 responds to are located in different levels of a CNN (color and orientation are lower levels and extracted from early layers while shapes are higher layer levels) we propose to approximate V4 region by taking as input the different layers of a pre-trained CNN (aiming to extract color, orientation, shape, etc.) and directly pass them to our adaptation module. 
    It then becomes obvious that this learned adaptation module directly wired to the V4 visual features simulated by the frozen pre-trained CNN is our model of the Primate IT cortex we propose to learn in a contrastive setting. 
}

\subsection{Adaptation function learning}

We look for an adaptation function able to bring closer the left embeddings ($d_{left}$) to the right ones ($d_{right}$) of the corresponding pairs closer than any other data. 
We model our adaptation module as the multiplication between a (learnable) weight matrix $W$ and input features $d_{left}$, $d_{right}$ followed by a ReLU activation. 
We will refer to the adaptation of the $d_{left}$, $d_{right}$ features as $a_{left}$, $a_{right}$ (Fig.~\ref{fig:pipeline}~(b)).

Previous works used contrastive or triplet loss \cite{chopra2005learning, hoffer2015deep} when learning from pairs or triplets.
To avoid sampling issue, as the whole pre-processed embeddings are light in memory, we build a similarity matrix by measuring similarity of each left image to each right image with the $\phi(\cdot, \cdot)$ similarity function.
The corresponding GT is the identity matrix.
We can thus learn to classify by using a softmax activation, followed by a Cross-Entropy loss function $\loss$.
We learn to find the right image from left queries and left images from right queries simultaneously, by averaging the directed $\loss_\text{left to right}$ and $\loss_\text{right to left}$ losses as in Eq.~\ref{eq:loss}.

\begin{equation}
    \begin{aligned}
        \loss_\text{left to right} = & \text{ CrossEntropy}(\text{softmax}(\phi(a_{left}, a_{right}) \cdot \sigma), \identity) \\
        \loss_\text{right to left} = & \text{ CrossEntropy}(\text{softmax}(\phi(a_{right}, a_{left}) \cdot \sigma), \identity) \\
        \loss = & \text{ }(\loss_\text{left to right} + \loss_\text{right to left}) / 2
    \end{aligned}
    \label{eq:loss}
\end{equation}
We used a parameter $\sigma$ in Eq.~\ref{eq:loss}, often referred as a temperature parameter in the literature.
This specific parameter $\sigma$ has already been used with success in previous works such as in \cite{Wu2018}.
In our case, by using a large $\sigma$, we are able to peak the softmax distribution to near binary values.
This can be viewed as a form of regularization.
Let us consider the case where the pair is successfully matched, without $\sigma$ the loss would still continue to try to bring the adapted embeddings even closer while they already are the best matches.
But, when a large $\sigma$ is used, the correct best match will have a value close to 1 not penalizing anymore the loss for right classification and thus not trying to continue to learn on adapted embeddings which are already optimal.
We show the positive influence on learning and generalization by using a large $\sigma$ in Table.~\ref{table_scores}.

\section{Experimental study}
\label{sec:evaluation}

\subsection{TTL Dataset}
We considered in our experimental study the \textit{Totally Looks Like} (\textit{TLL}) image dataset introduced earlier. 
It is composed of 6016 image pairs, perceptually similar but semantically unrelated (some illustrative samples are provided in Fig.~\ref{fig:ttl_dataset}).
The images come from very different domains such as photography, cartoons, paintings, sketches, logos, etc.

The dataset can be split into two sub-datasets: 
one of 1817 pairs containing only well centered faces (noted \ttlfaces{}) and one of 4199 pairs of images captured from the wild (noted \ttlobj{}). 
Since \cite{Sadovnik_2018_CVPR_Workshops} already focused on the particular case of facial similarity with data richer than pairs on a bigger dataset, we will focus on the more general case of the 4199 remaining pairs.
To subtract the \ttlfaces{} subset from the whole \textit{TLL}, we labeled as face-pairs only pairs where faces were detected in both right and left images with a Haar Cascade classifier.
In this context, the remaining set (\ttlobj{}) still contains pairs where faces are compared to other animals, objects, paintings, etc. due to strange facial expressions or other features.

\subsection{Protocol}

\begin{figure}[t]
%\begin{minipage}[b]{1.0\linewidth}
  \centering
%   \centerline{\includegraphics[trim=15 10 50 50,clip,width=0.8\linewidth]{imgs/new_scores.png}}
    \centerline{\includegraphics[width=1\linewidth]{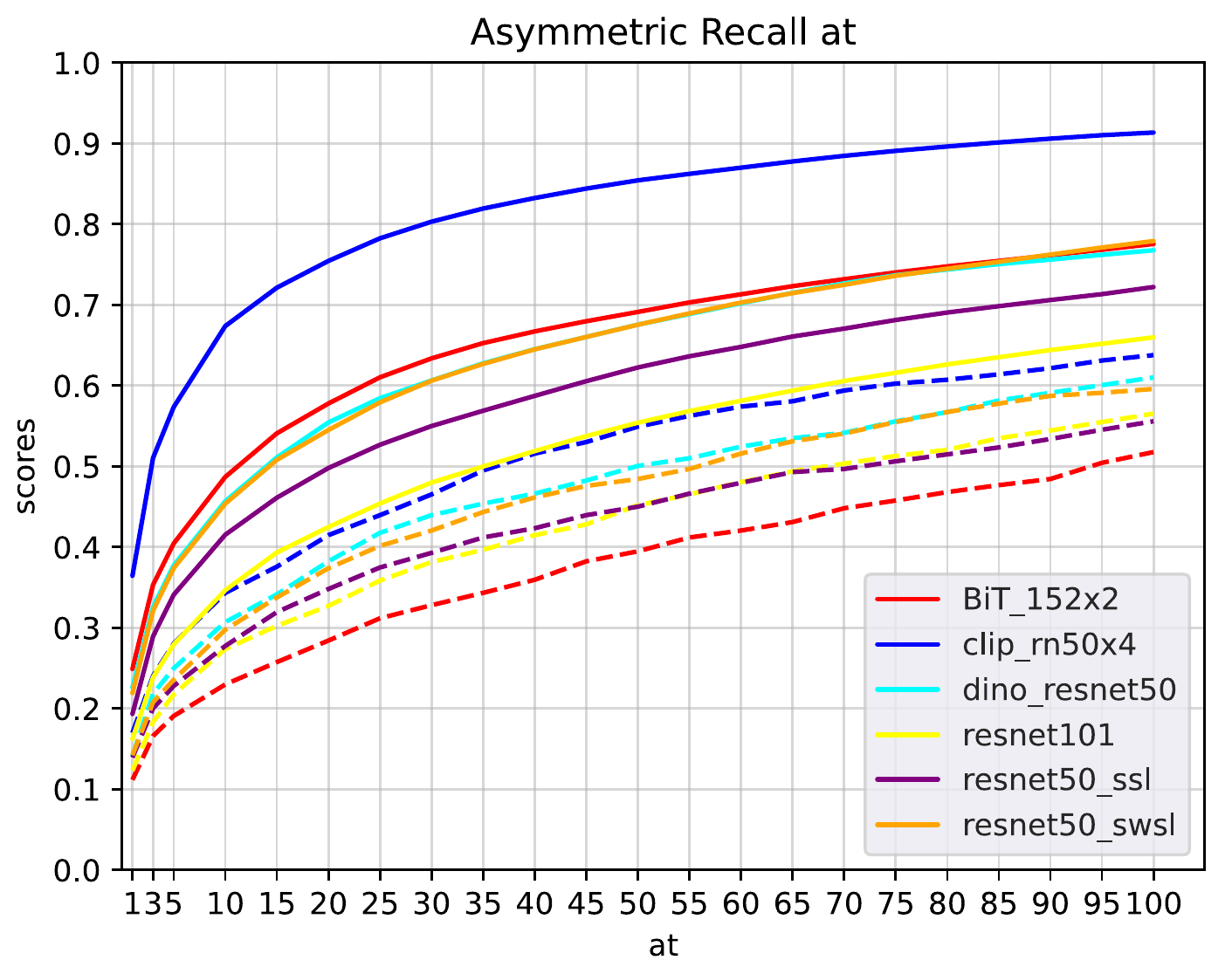}}
  \caption{Comparison of \textit{Asymmetric Recall curves at different ranks between baselines with $\sigma=15$}. Dashed line correspond to performances before adaptation, plain ones correspond to the adapted ones.}
%\end{minipage}
\label{fig:asymmetric_recall_256to1024_20runs_all-temperatures}
\end{figure}

\begin{figure}[t]
%\begin{minipage}[b]{1.0\linewidth}
  \centering
  %\centerline{\includegraphics[width=0.9\linewidth]{imgs/test_examples.png}}
  \centerline{\includegraphics[width=0.8\linewidth]{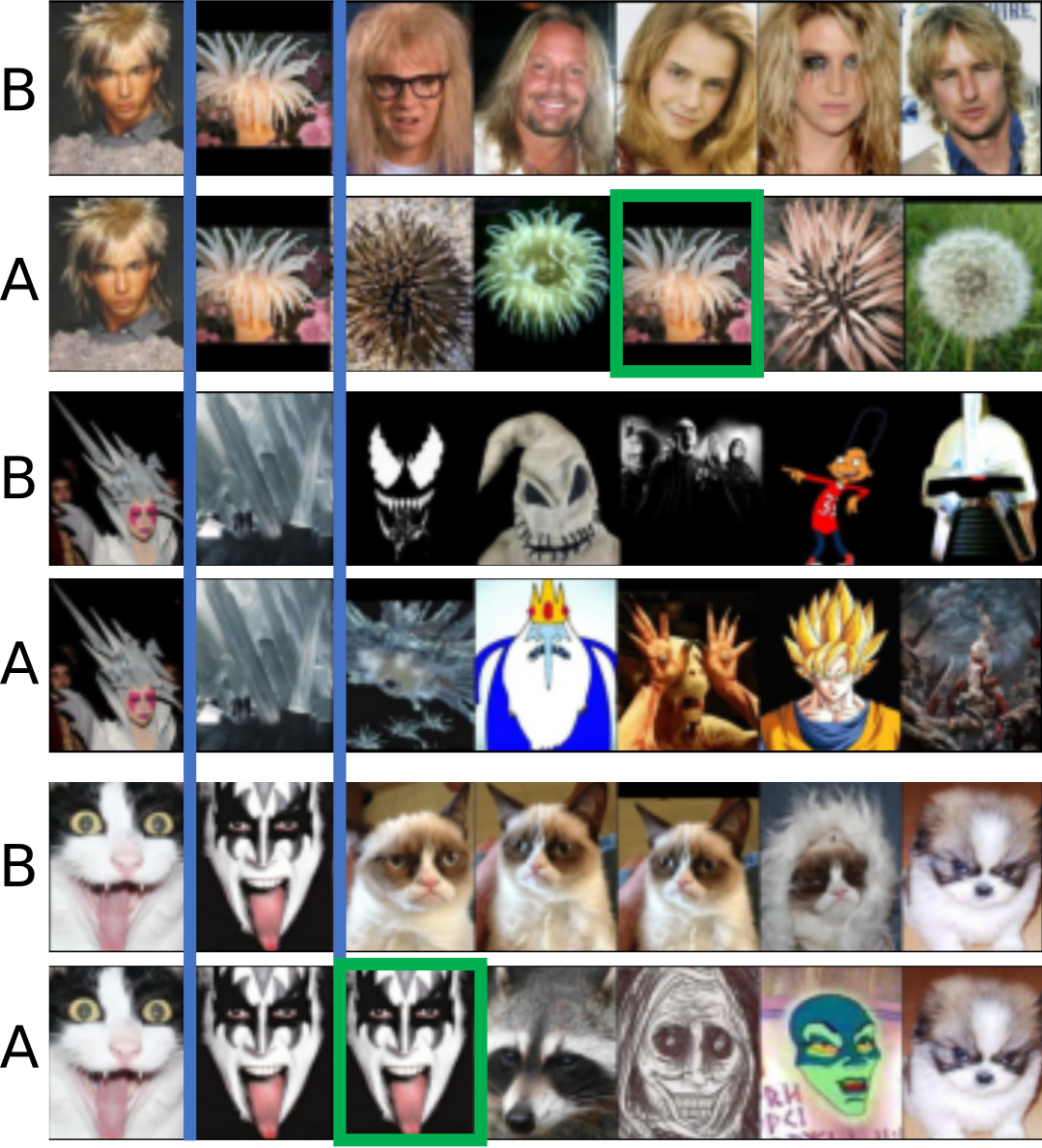}}
  \caption{Results before (B) and after (A) adaptation. 
  The first column corresponds to the query, the second to the GT and the other images correspond to the returned candidates.}
%\end{minipage}
\label{fig:test_examples}
\end{figure}

For each experiment, the \ttlobj{} dataset of 4199 pairs was divided into 75-25 train-test sets.
We did not use a validation set due to the non significativness of the scores obtained on too small validation or testing bases, furthermore, if we decrease the training set size, the exhaustivity of the different ways that human compare images is too partial.

We used a PCA with 256 dimensions to compress the pre-computed 
%\textit{ResNet50} 
embeddings into $d_{left}$, $d_{right}$.
Experimentally, we found that 256 was able to remove the noise reducing both, the input space and overfitting of our adaptation. 
Thus we will use those compressed vectors both as a second baseline and as inputs of our method.
We used then an adaptation vector size of 1024, thus the matrix $W$ is sized from 256 to 1024. 
{As we do not have a validation set, we experimentally set the number of epochs to 150 and we selected the optimal number of epochs by bootstrapping for each model.} 
This provides a good compromise between low overfitting and the best possible model if we had an oracle to select it.

Due to the small number of examples in the testing base and the stochasticity induced by the random split and matrix weights initialization, we run our experiments 20 times with different splits and matrix initializations.

\subsection{Evaluation and results}

As discussed previously, human similarity judgment has been found to be asymmetric \cite{tversky1977features}.
Since we do not have the asymmetry indications in the \textit{TLL} dataset, we propose to evaluate the results in an optimistic way: if the pairing is found in one direction or another, the pairing is considered as successful. 
We refer to this optimistic recall as \textit{Asymmetric Recall} ($aR$).

Results are reported on Fig.~\ref{fig:asymmetric_recall_256to1024_20runs_all-temperatures} and 
Tab.~\ref{table_scores}.
We computed the \textit{Asymmetric Recall} on 20 random tests splits and display the mean scores surrounded by two standard deviations. 
From these results, we observe the strong influence of the temperature $\sigma$ on the results.
Our method is compared to the baselines (with and without) dimension reduction.
%{(Faudra changer avec les vrais nouveaux scores) 
We can see a $2.25\times$, $3.55\times$ and $4.53\times$ average improvement for $aR@\{1, 5, 20\}$ by only using the \textit{Clip\_RN50$\times$4} features. 

We provide a few qualitative results in Fig.~\ref{fig:test_examples} from the test set.
The second case shows that none of the previous neither our method is able to find the right image, however our method was able to capture the sharp hairstyle and glacial aspect of the picture.
The first and last ones show the limitation of purely semantic features for visual retrieval, while our method leads to more diverse results.
On the last example, it can be noticed that while a semantically biased features extractor will firstly output cats when querying a cat, our method finds image from very different classes, still sharing connection on the dark eyes surrounding.

\subsubsection{Comparative study}

% \hala
{Datasets scales and contents influence the results of each learning approach, which made us question the importance of the pre-trained network chosen for the features extraction module.}
In the Table~\ref{table_scores} in annex
we compare Residual networks of different depths: ResNet18, ResNet50, ResNet101, ResNet152, \cite{He2016}, 
the self-supervised DINO framework pre-trained with ResNet50 \cite{dino2021}, 
FaceNet pre-trained on VGGFace2 \cite{Schroff2015}, BiT-M of different sizes: ResNet-50x1,  ResNet-50x3,  ResNet-101x1,  ResNet-101x3,  and  ResNet-152x4 \cite{BiT}, Barlow Twins Self-Supervised model pre-trained on ResNet50 \cite{barlow} and CLIP, a framework that works on learning visual concepts from natural language supervision \cite{clip2021}.

{In \cite{zhuang2021unsupervised}, they found that the unsupervised models as well as self-supervised ones were the best at modelling and approximating the ventral visual stream, though their similarity task is different from ours. 
We compared different learning types to confirm that self-supervised models are the best ones for approximating the visual ventral system, which is coherent with their findings.}

\section{Discussions}
In this article, we focus on the specific task of learning visual image similarities. Inspired by previous works on neural features adaptation to psycho-cognitive representations,
we proposed a method to adapt semantic neural representations to visual ones. In average, our method improves the
retrieval score up to 4.53×. 
We observed as well a qualitative improvement on the returned candidates. In addition,
we saw that the CLIP framework was the best in our case,
and we assume that it is due to two things: (1) First, such model has been trained on
a dataset which consists of 400M images gathered from
the internet; (2) Second, because
of the robustness of the CLIP model and its ability to generalize the concepts as well as its flexibility when it comes
to semantic bias.
Despite those improvements, some retrieval results are still
not meaningful. Accordingly to \cite{German2020}, in contrary to humans, neural features may be less suited to achieve viewpoints invariance. Moreover, as for several pairs the shared
visual similarity relies on few pixels, this problem could
be tackled in a fine-grained framework. Since pairs in the
TLL dataset are not equally sampled from the meta-features
of an image (color, texture, layout, etc.), identifying those
different meta-features used by human to judge about the
visual similarity between images could help to re-balance
learning.
Further investigations are ongoing.

\bibliographystyle{IEEEbib}
\bibliography{citations}

\newpage

\begin{landscape}
\begin{appendices}
\section{}

\vspace{1.5cm}

\begin{table}[h]
\centering
\caption{Asymmetric Recall scores of each model before and after adaptation (our method). 
We only extracted features from one of the last layer in models marked with a $ * $.
Scores are averaged on 20 runs and reported with the standard deviation.
}
\resizebox{\linewidth}{!}{ 
\begin{tabular}{llllllllll}
%\toprule
\hline
Model &          Learning type &      Dataset & Number of images (M) & Concatenated Size & PCA Variance Sum & Concatenated Score & PCA Score & Our aR@1 $\sigma=1$ & Our aR@1 $\sigma=15$ \\
\hline
%\midrule

barlow                &               self-supervised &  ImageNet-1k &             1.28 &              3840 &           0.7555 &            0.1389 $\pm 0.0106$ &           0.1357 &            0.1781 $\pm 0.0041$ &             0.2065 $\pm 0.0045$ \\

deits16 *               &               self-supervised &  ImageNet-1k &             1.28 &               384 &           0.9352 &          0.1392 $\pm 0.0107$ &           0.1362 &            0.1456 $\pm 0.0044$ &     0.1826 $\pm 0.0044$                 \\
deits8 *               &               self-supervised &  ImageNet-1k &             1.28 &               384 &           0.9451 &           0.149~~ $\pm 0.0109$ &           0.1477 &            0.1623 $\pm 0.0057$  &       0.2035 $\pm 0.0057$             \\
dino\_resnet50          &               self-supervised &  ImageNet-1k &             1.28 &              3840 &           0.7732 &          0.136~~ $\pm 0.0089$ &           0.1288 &            0.1947 $\pm 0.0034$ &             \textbf{0.2435 $\pm 0.0040$} \\
dino\_vitb16 *            &               self-supervised &  ImageNet-1k &             1.28 &               768 &           0.8505 &        0.1456 $\pm 0.0100$ &           0.1394 &            0.1745 $\pm 0.0039$  &          0.2009 $\pm 0.0048$         \\
dino\_vitb8 *            &               self-supervised &  ImageNet-1k &             1.28 &               768 &           0.8199 &         0.1334 $\pm 0.0122$ &           0.1314 &            0.1763 $\pm 0.0041$  &          0.2012    $\pm 0.0047$      \\

\hline

facenet *               &                    supervised &     VGGFace2 &             3.31 &              1792 &           0.9722 &          0.0533 $\pm 0.0050$ &           0.0611 &            0.0739 $\pm 0.0035$ &             0.1173 $\pm 0.0038$ \\
alexnet *               &                    supervised &  ImageNet-1k &             1.28 &               256 &           1.0000 &          0.1023 $\pm 0.0080$ &           0.1041 &            0.1061 $\pm 0.0043$ &           0.1321   $\pm 0.0045$ \\
resnet18               &                    supervised &  ImageNet-1k &             1.28 &               960 &           0.9118 &           0.1283 $\pm 0.0097$ &           0.1259 &            0.1323 $\pm 0.0035$ &             0.1684 $\pm 0.0044$ \\
resnet34               &                    supervised &  ImageNet-1k &             1.28 &               960 &           0.9011 &           0.1236 $\pm 0.0091$ &           0.1304 &            0.1426 $\pm 0.0037$ &        0.1469     $\pm 0.0047$ \\
resnet50               &                    supervised &  ImageNet-1k &             1.28 &              3840 &           0.8832 &           0.1211 $\pm 0.0073$ &           0.1258 &            0.1340 $\pm 0.0051$ &             \textbf{0.1913 $\pm 0.0034$} \\
resnet101              &                    supervised &  ImageNet-1k &             1.28 &              3840 &           0.8974 &           0.1209 $\pm 0.0073$ &           0.1212 &            0.1557 $\pm 0.0042$ &             0.1686 $\pm 0.0046$ \\
resnet152              &                    supervised &  ImageNet-1k &             1.28 &              3840 &           0.8971 &           0.125~~ $\pm 0.0094$ &           0.1214 &            0.1472 $\pm 0.0044$ &             0.1892 $\pm 0.0040$ \\

efficientnet\_b0 *        &                    supervised &  ImageNet-1k &             1.28 &              1280 &           0.7725 &        0.1164 $\pm 0.0093$ &           0.1149 &            0.1232 $\pm 0.0039$ &          0.1552  $\pm 0.0032$ \\
efficientnet\_b1 *       &                    supervised &  ImageNet-1k &             1.28 &              1280 &           0.7630 &         0.1243 $\pm 0.0067$ &           0.1242 &            0.1384 $\pm 0.0033$ &       0.1387      $\pm 0.0034$  \\
efficientnet\_b2 *       &                    supervised &  ImageNet-1k &             1.28 &              1408 &           0.7423 &         0.1165 $\pm 0.0071$ &           0.1170 &            0.1205 $\pm 0.0029$  &         0.1602    $\pm 0.0042$ \\
efficientnet\_b3 *       &                    supervised &  ImageNet-1k &             1.28 &              1536 &           0.7188 &         0.1162 $\pm 0.0090$ &           0.1126 &            0.1205 $\pm 0.0033$ &         0.1410    $\pm 0.0046$ \\

ViT *                    &                    supervised &  ImageNet-1k &             1.28 &               768 &           0.8001 &         0.121~~ $\pm 0.0084$ &           0.1173 &            0.1377 $\pm 0.0032$ &          0.1570   $\pm 0.0035$ \\

\hline

BiT-M-R50x1                &                    supervised &     ImageNet-21k &             14.19 &              3840 &           0.9212 &  0.1165 $\pm 0.0076$ &           0.1142 &            0.1722 $\pm 0.0031$ &             0.2332 $\pm 0.0042$ \\

BiT-M-R50x3                &                    supervised &     ImageNet-21k &             14.19 &              11520 &           0.8768 & 0.1165 $\pm 0.0099$ &           0.1143 &            0.2049 $\pm 0.0046$ &             0.2659 $\pm 0.0052$ \\

BiT-M-R101x1                &                    supervised &     ImageNet-21k &             14.19 &              3840 &           0.9174 & 0.1114 $\pm 0.0087$ &           0.1054 &            0.2030 $\pm 0.0051$ &             0.2509 $\pm 0.0055$ \\

BiT-M-R101x3                &                    supervised &     ImageNet-21k &             14.19 &              11520 &       0.8789 &    0.1099 $\pm 0.0087$ &           0.1075 &            0.1800 $\pm 0.0040$ &             0.2437 $\pm 0.0039$ \\

BiT-M-R152x2                &                    supervised &     ImageNet-21k &             14.19 &              7680 &           0.8877 & 0.1171 $\pm 0.0058$  &           0.1101 &            0.2215 $\pm 0.0060$ &             \textbf{0.2732 $\pm 0.0049$} \\

BiT-M-R152x4                &                    supervised &     ImageNet-21k &             14.19 &              15360 &       0.8033 &    0.1174 $\pm 0.0102$ &           0.1143 &            0.2145 $\pm 0.0024$ &             0.2540 $\pm 0.0052$ \\

\hline

resnet50\_ssl           &               semi-supervised &  ImageNet-1k &             1.28 &              3840 &           0.7973 &          0.1396 $\pm 0.0075$ &           0.1401 &            0.1883 $\pm 0.0039$ &         \textbf{0.1995 $\pm 0.0051$}  \\
resnext101\_32x8d\_ssl   &               semi-supervised &  ImageNet-1k &             1.28 &              3840 &           0.8539 &         0.1255 $\pm 0.0089$ &           0.1276 &            0.1589 $\pm 0.0033$ &         0.1550 $\pm 0.0040$ \\

\hline

resnet50\_swsl          &        weakly semi-supervised &  ImageNet-1k &             1.28 &              3840 &           0.7912 &          0.1548 $\pm 0.0099$ &           0.1506 &            0.1929 $\pm 0.0035$ &         \textbf{0.2191 $\pm 0.0038$} \\ 
resnext101\_32x16d\_swsl &        weakly semi-supervised &  ImageNet-1k &             1.28 &              3840 &           0.8729 &         0.1213 $\pm 0.0068$ &           0.1270 &            0.1414 $\pm 0.0031$ &         0.1986 $\pm 0.0049$    \\
resnext101\_32x8d\_swsl  &        weakly semi-supervised &  ImageNet-1k &             1.28 &              3840 &           0.8702 &         0.1295 $\pm 0.0067$ &           0.1297 &            0.1544 $\pm 0.0029$ &         0.2005 $\pm 0.0041$ \\

\hline

resnext101\_32x8d\_wsl   &             weakly supervised &  ImageNet-1k &             1.28 &              3840 &           0.7861 &         0.1297 $\pm 0.0094$ &           0.1192 &            0.1642 $\pm 0.0035$ &         0.1917  $\pm 0.0041$   \\
resnext101\_32x16d\_wsl  &             weakly supervised &  ImageNet-1k &             1.28 &              3840 &           0.7731 &         0.1292 $\pm 0.0079$ &           0.1204 &            0.1670 $\pm 0.0035$ &          0.2061 $\pm 0.0047$  \\
resnext101\_32x32d\_wsl  &             weakly supervised &  ImageNet-1k &             1.28 &              3840 &           0.7651 &         0.1406 $\pm 0.0091$ &           0.1323 &            0.1639 $\pm 0.0047$ &         0.2295 $\pm 0.0052$  \\
resnext101\_32x48d\_wsl  &             weakly supervised &  ImageNet-1k &             1.28 &              3840 &           0.7629 &         0.1458 $\pm 0.0085$ &           0.1419 &            0.1696 $\pm 0.0037$ &      \textbf{0.2318 $\pm 0.0046$}  \\

\hline

clip\_rn101             &  natural language supervision &     Internet &            400.0 &              4352 &           0.8426 &          0.1674 $\pm 0.0104$ &           0.1658 &            0.2814 $\pm 0.0050$ &             0.3825 $\pm 0.0046$ \\
clip\_rn50              &  natural language supervision &     Internet &            400.0 &              4864 &           0.8679 &          0.1575 $\pm 0.0102$ &           0.1606 &            0.2834 $\pm 0.0059$ &             0.3607 $\pm 0.0047$ \\
clip\_rn50x4            &  natural language supervision &     Internet &            400.0 &              5440 &           0.8199 &          0.1743 $\pm 0.0070$ &           0.1700 &            0.3282 $\pm 0.0041$ &             \textbf{\underline{0.3939 $\pm 0.0071$}} \\
clip\_vit32\_b *          &  natural language supervision &     Internet &            400.0 &               768 &           0.8600 &        0.1714 $\pm 0.0074$ &           0.1797 &            0.2655 $\pm 0.0068$  &             0.3540 $\pm 0.0046$ \\

\hline

\end{tabular}
}
\label{table_scores}
\end{table}

\end{appendices}
\end{landscape}

\end{document}